\def\year{2017}\relax
\documentclass[letterpaper]{article}
\usepackage{aaai17}
\usepackage{times}
\usepackage{helvet}
\usepackage{courier}
\usepackage{graphicx}
\usepackage[]{amsmath}
\usepackage{amsfonts}
\usepackage{subcaption}
\begin{document}
%
\title{Toward Implicit Sample Noise Modeling:\\ Deviation-driven Matrix Factorization}
\author{Guang-He Lee\\
Dept. of Computer Science\\
National Taiwan Univ., Taiwan\\
r04922045@csie.ntu.edu.tw\\
\And
Shao-Wen Yang\\
Intel Labs\\
Intel Corporation\\
shao-wen.yang@intel.com\\
\And
Shou-De Lin\\
Dept. of Computer Science\\
National Taiwan Univ., Taiwan\\
sdlin@csie.ntu.edu.tw\\
}

\maketitle
\begin{abstract}
The objective function of a matrix factorization model usually aims to minimize the average of a regression error contributed by each element. However, given the existence of stochastic noises, the implicit deviations of sample data from their true values are almost surely diverse, which makes each data point not equally suitable for fitting a model. In this case, simply averaging the cost among data in the objective function is not ideal. Intuitively we would like to emphasize more on the reliable instances (i.e., those contain smaller noise) while training a model. Motivated by such observation, we derive our formula from a theoretical framework for optimal weighting under heteroscedastic noise distribution. Specifically, by modeling and learning the deviation of data, we design a novel matrix factorization model. Our model has two advantages. First, it jointly learns the deviation and conducts dynamic reweighting of instances, allowing the model to converge to a better solution. Second, during learning the deviated instances are assigned lower weights, which leads to faster convergence since the model does not need to overfit the noise. The experiments are conducted in clean recommendation and noisy sensor datasets to test the effectiveness of the model in various scenarios. The results show that our model outperforms the state-of-the-art factorization and deep learning models in both accuracy and efficiency.
\end{abstract}

\section{Introduction}

Factorization models have been widely and successfully applied to tasks such as missing data imputation and user-item recommendation. The objective function of a factorization model usually aims to minimize an average of a regression cost contributed by each data point \cite{koren2009matrix}. However, in general, data can be noisy, and each data point may be contaminated by different level of noise. As long as the noise comes from certain continuous probability distribution, almost surely the sample data will deviate diversely from their expected (noiseless) values. Namely, the existence of noise will cause certain instances (i.e., those with larger noise) to be less trust-able than others.

Given certain level of trustworthiness for each data point, the instance-weighting strategy is widely adopted \cite{koren2009matrix}. However, such weighting scheme has its limitation since it requires prior knowledge about the trustworthiness of each data point. A related study may be anomaly/outlier detection \cite{xiong2011direct}, whose goal is to identify a few (salient) noisy instances from a majority of noiseless data. Our model, on the other hand, assumes the noise is implicitly embedded in every data point, while some contain more and some less. This is particularly true for sensor data where measurement errors always exist and different sensor has different degree of errors. Therefore, we argue that a learning model should consider different level of noise in each data point instead of simply detect and remove noisy data from the training set. 

\begin{figure} 
\centering
\includegraphics[width=1\linewidth]{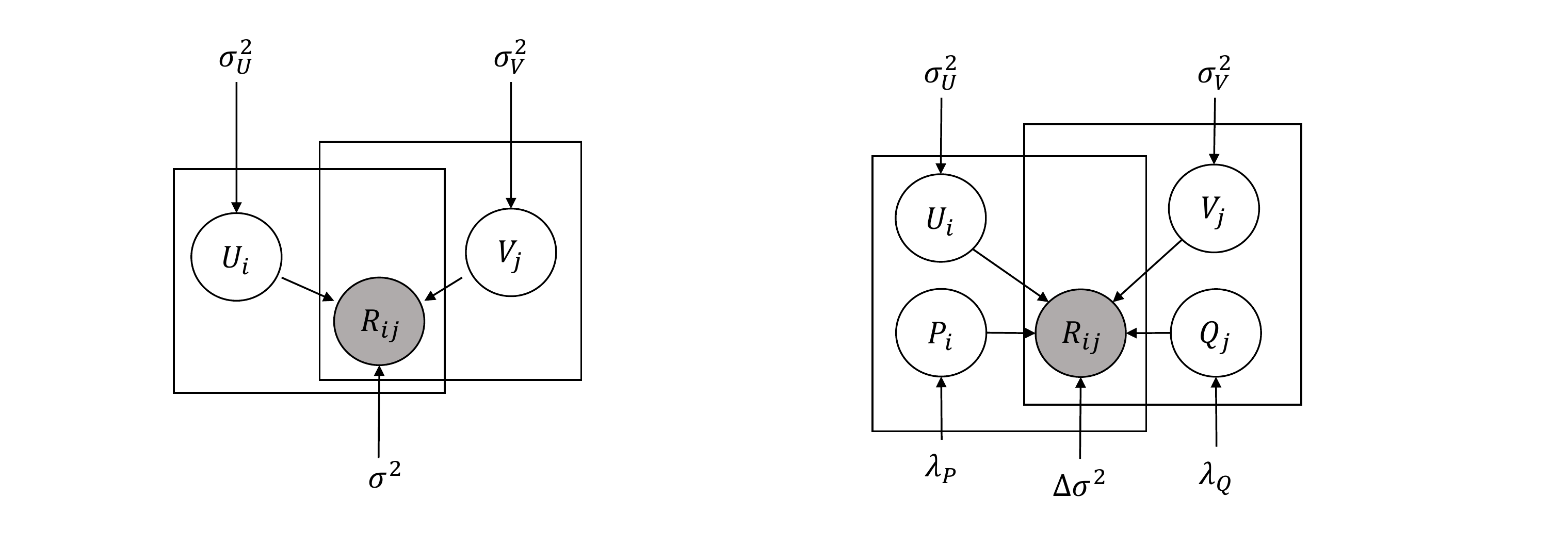}
\caption{Graphical model representation of probabilistic matrix factorization (PMF) on the left vs. deviation-driven matrix factorization (DMF) on the right. The factorization $P_i$, $Q_j$ and $\Delta\sigma^2$ generates instance-dependent variance $\bar{\sigma}^2_{ij}$.
}
\label{fig:pgm}
\end{figure}



To address the above issues, we propose a \emph{deviation-driven modeling} framework on matrix factorization (MF) to model and learn the deviation, i.e., implicit sample noise. Note that we emphasize the implicit and ubiquitous nature of sample noise, which is possibly unobservable and universally existent for data with stochastic noise. As a result, though the concept of modeling noise is closely related to robust methods \cite{meng2013robust}, our model should be treated as a general method.

Specifically, as the concept of deviation is similar to heteroscedasticity, we derive a deviation-driven loss function from a theoretical framework for optimal weighting of weighted linear regression under heteroscedastic noise distribution. By learning the deviation from data, we relieve one less-realistic assumption of the theory that the level of noise is known. The resulting formulation is equivalent to a Gaussian likelihood with heteroscedastic variance. The formulation exhibits two advantages. First, it jointly learns the deviation and conducts dynamic reweighting of instances, allowing the model to converge to a better solution. Second, during learning the deviated instances are assigned lower weights, which leads to faster convergence since the model does not need to overfit the noise.


We then apply the deviation-driven loss to MF as Deviation-driven Matrix Factorization (DMF). In contrast to traditional sparse noise structure that assumes only limited amount of noise, a novel low-rank structure for modeling the deviation, i.e., variance in the Gaussian likelihood, is adopted to model the ubiquitous nature of deviation and combat overfitting. The illustration of DMF vs. traditional Probabilistic Matrix Factorization is shown in Figure \ref{fig:pgm}. 

Several experiments are conducted in both \emph{noisy} sensor and \emph{clean} recommendation datasets. The "clean" datasets are especially important to verify the existence of implicit noise, which is not directly observable and can be implied from the superiority of noise modeling over typical methods. In both scenario, our model demonstrates better accuracy and faster training than the state-of-the-art solutions.

\section{The Deviation-driven Model}


\subsection{The Diverse Deviation Phenomenon}

\begin{figure}
\centering
\includegraphics[width=1\linewidth]{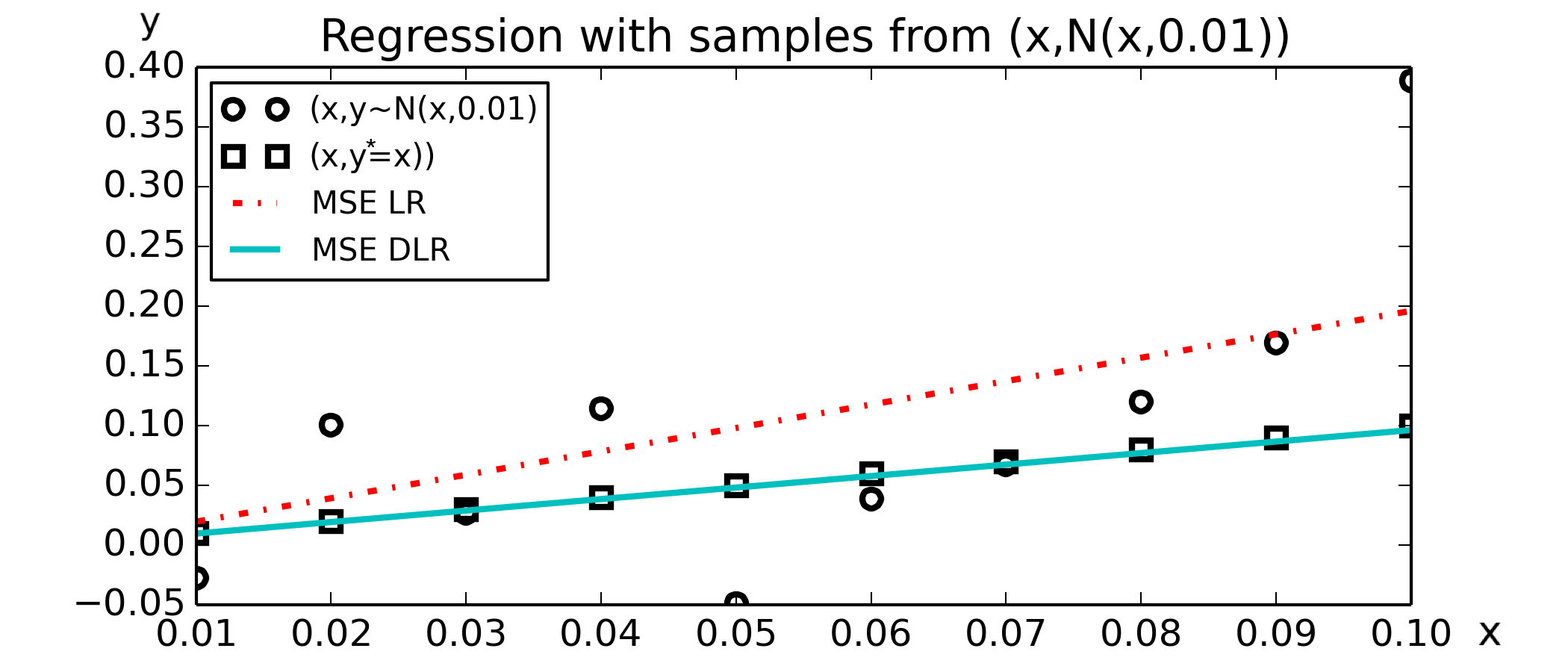}
\caption{Data points drawn from $(x,y \sim \mathcal{N}(x,0.01))$ as circle markers. The expected data points of the distribution $(x,y^*=x)$ are denoted as square markers. Despite a shared variance, the deviations for data points from their mean are diverse. In this setting, a deviation-driven linear regression (DLR) fits the distribution much better than a linear regression (LR) model.}
\label{fig:wlr}
\end{figure}

In this work, we assume each observed data instance $(x_i,y_i)$ comes from the addition of clean data instance $(x_i,y_i^*)$ and sample noise $\epsilon_i$, \emph{sampled} from a continuous random variable $\varepsilon_i$. Besides, the expectation of noise is assumed to be zero. Formally speaking, denoting $\dot{\sim}$ as a sampling process, we have
\begin{equation}
(x_i,y_i=y_i^*+\epsilon_i), \epsilon_i \dot{\sim} \varepsilon_i, E[\varepsilon_i]=0.
\end{equation}
For example, in Figure \ref{fig:wlr}, we conduct a random experiment to generate sample data points with $y_i^*=x$ and $\varepsilon_i \sim \mathcal{N}(0,0.01)$ as circle markers, and correspondingly expected data points as square marker. Even with homoscedastic noise, the sample data still deviated from their mean diversely. Theoretically, given \emph{any} two continuous noises $\varepsilon_i$ and $\varepsilon_j$ with probability density functions $p(x)$ and $q(x)$, it is almost surely that the two noises are diverse, as $Pr(\varepsilon_i \neq \varepsilon_j) = \int_{-\infty}^\infty p(x)(1-\int_{x}^x q(y)\mathrm{d}y)\mathrm{d}x = \int_{-\infty}^\infty p(x)\mathrm{d}x = 1.$ Accordingly, as each data point brings different level of confidence for describing the original distribution, to conduct a fair treatment for data, it is important to consider the diverse deviation for individual data point. 

To address the above issue, an intuitive method is to conduct a weighting scheme for each data point. We begin our derivation from a theoretical framework on a linear model, weighted linear regression (WLR), which solves the following optimization task:
\begin{equation}
w = \operatorname{argmin}_{\bar{w}}  \frac{1}{N}\sum_{i=1}^{N} \frac{(y_i-\bar{w}^Tx_i)^2}{W_i},
\end{equation}
where $W_i$ is the weighting for instance $(x_i,y_i)$ and $w$ is the linear model to optimize. Given that data are generated by a linear model $w^*$ as $y_i=y^*_i+\epsilon_i=(w^*)^Tx_i+\epsilon_i$, it is reasonable to minimize $E[(w-w^*)^2]=Var(w)$ with respect to $W_i$, since $E[y_i]=y_i^*$ and thus $E[w]=w^*$. Accordingly, from the \emph{generalized Gauss-Markov theorem} \cite{GAUSSMARKOV}, the optimal $W_i$ is $Var(\varepsilon_i)$, the variance of noise.

Since random variable $\varepsilon_i$ is unobservable in \emph{sampled data}, we make an unrealistic assumption that \emph{sample deviation} $\epsilon_i$ can be observed. Accordingly, we can use the square of sample deviation $\epsilon_i^2$ to approximate $Var(\varepsilon_i)$ as (\ref{model_noise}), which we call it the deviation-driven (weighted) linear regression (DLR).

\begin{equation}
\label{model_noise}
W_i = Var(\varepsilon_i) = E[(\varepsilon_i-E[\varepsilon_i])^2] = E[\varepsilon_i^2] \approx 
\epsilon_i^2.
\end{equation}

In Figure 2, we compare linear regression (LR) with DLR. The superior performance of DLR is clear for fitting the clean (expected) data. The simple experiment demonstrates the promising efficacy of incorporating deviation, which motivates modeling deviation in more realistic scenario.

\subsection{Deviation-driven Squared Error} 
 To extend DLR for MF on a matrix $R$ with size $N_1 \times N_2$, we can correspond the label $y_i$ with matrix element $R_{ij}$ and the label prediction $w^Tx_i$ with matrix \emph{prediction} $\bar{R}_{ij}$. However, the weight $W_i=\epsilon_i^2$ is unknown in practice. Thus, certain strategies to learn the deviation are needed. Here we propose to use a positive \emph{estimation} $\bar{\sigma}^2_{ij}$ to model the deviation $\epsilon_i^2$, together with a deviation regularizer $\Psi$ on the value of $\bar{\sigma}^2_{ij}$ to prevent it from growing unlimitedly, since a larger $\bar{\sigma}^2_{ij}$ leads to smaller cost for data point $R_{ij}$. The final deviation-driven squared error objective function is shown in (\ref{lrefor_object}). 

\begin{equation}
\operatorname{min} \frac{1}{N_1N_2}\sum_{i=1}^{N_1}\sum_{j=1}^{N_2} (\frac{(R_{ij}-\bar{R}_{ij})^2}{\bar{\sigma}^2_{ij}} + \Psi(\bar{\sigma}^2_{ij})) \label{lrefor_object}.
\end{equation}

Since there are two variables to estimate, $\bar{R}_{ij}$ and $\bar{\sigma}^2_{ij}$, we will use (main) prediction to describe $\bar{R}_{ij}$, and deviation estimation to describe $\bar{\sigma}^2_{ij}$ throughout the paper. 

If adopting a natural log deviation regularizer $\Psi=\ln$, it is not hard to see that the formula (\ref{lrefor_object}) is equivalent to the maximization of Gaussian likelihood as,
\begin{equation}
\operatorname{max} \prod_{i=1}^{N_1}\prod_{j=1}^{N_2} \frac{1}{\bar{\sigma}_{ij} \sqrt[]{2\pi}}e^{-\frac{(R_{ij}-\bar{R}_{ij})^2}{2\bar{\sigma}_{ij}^2}}. \label{gauss2}
\end{equation}
The regularization term $\Psi(\bar{\sigma}^2_{ij})=\ln(\bar{\sigma}^2_{ij})$ in (\ref{lrefor_object}) corresponds to the $\bar{\sigma}_{ij}$ in the denominator in (\ref{gauss2}).

The representation provides a physical meaning of deviation as the variance, which is the theoretically optimal weighting in the \emph{generalized Gauss-Markov theorem} \cite{GAUSSMARKOV}, in a Gaussian likelihood. Moreover, contrasting most conventional modeling for Gaussian likelihood with \emph{shared} variance \cite{agarwal2010flda,mnih2007probabilistic,salakhutdinov2008bayesian}, the deviation-driven modeling represents Gaussian likelihood in a per-instance variance manner.


\subsection{Deviation-driven Matrix Factorization}

\subsubsection{Matrix Factorization}

Given a matrix $R$, MF reconstruct it as matrix $\bar{R}$ by low rank matrices $U$ and $V$ with latent dimension $D$. A widely used variant \cite{chen2011linear,koren2009matrix} is to add some bias and scalar terms to the factorization as follows:

\begin{align}
\bar{R}_{ij} = U_i^TV_j + u_i + v_j + \mu. 
\label{b_MF}
\end{align}

$u_i$ and $v_j$ are biases, which are scalars to be learned, for user $i$ and item $j$, respectively. $\mu$ is the average value for observed elements of $R$ in the training data. In the remainder of the paper, $u_i$, $v_j$, and $\mu$ are omitted for clarity purpose.

\subsubsection{Deviation-driven Modeling}

Here we propose the idea of Deviation-driven Matrix Factorization (DMF). Conventionally, the data can be modeled by a Gaussian likelihood with mean constructed by MF. 
To adapt MF to a deviation-driven model, we have to further model the variance in the Gaussian likelihood. The key concern lies in the choice of variance structure. Given the ubiquitous nature of noise, a sparse structure is unsuitable and it is necessary to build a prediction model for every element on the matrix. In order to regularize the capability of such prediction, a low-rank structure is desirable. Consequently, we can adopt another MF to construct the variance $\bar{\sigma}_{ij}^2$ by deviation latent factors, $P_i$ and $Q_j$.

Since variance has a non-negative constraint, we can impose a non-negative prior on deviation latent factors, thus the inner product of non-negative vectors is also non-negative. In addition, to avoid a zero value of $\bar{\sigma}_{ij}^2$ that leads to infinite objective value, we add a small positive value $\Delta \sigma^2$ as (\ref{l_sigma}).

\begin{equation}
\bar{\sigma}^2 _{ij} = P_i^TQ_j+\Delta \sigma^2 \label{l_sigma}.
\end{equation}
To impose a non-negative prior, we select exponential distribution. The usage of exponential prior leads to sparse solution \cite{gopalan2013scalable}, which is favorable for its efficiency in computation and storage. Accordingly, the prior distribution of deviation latent factor results in (\ref{exponential1}).

\begin{align}
\label{exponential1}
p(P_i|\lambda_{P}) = \mathcal{EXP}(P_i|\lambda_{P}I),\;  p(Q_j|\lambda_{Q}) = \mathcal{EXP}(Q_j|\lambda_{Q}I).
\end{align}


With Gaussian prior on mean latent factors, the mean prior and likelihood of DMF can be represented as the follows, with $I_{ij}$ being a binary indicator of the availability of $R_{ij}$ in training data.

\begin{align}
& p(U_i|\sigma_{U}^2) = \mathcal{N}(U_i|0,\sigma_{U}^2I),\; p(V_j|\sigma_{V}^2) = \mathcal{N}(V_j|0,\sigma_{V}^2I), \label{ptensor_uprior}\\
& p(R_{ij}|U_i,V_j,P_i,Q_j) = \mathcal{N}(R_{ij}|\bar{R}_{ij} ,\bar{\sigma}^2 _{ij})^{I_{ij}}.
\end{align}

The log posterior distribution of DMF is given as the following form,

\begin{align}
\label{log_posterior}
& - \sum_{i=1}^{N_1} \sum_{j=1}^{N_2}\frac{I_{ij}(R_{ij} - U_i^TV_j)^2}{2(P_i^TQ_j+\Delta \sigma^2)} - \sum_{i=1}^{N_1} (\frac{U_i^T U_i}{2\sigma_{U}^2}+\lambda_{P} P_i^T I) \nonumber\\
& - \sum_{j=1}^{N_2} (\frac{V_j^T V_j}{2\sigma_{V}^2}+\lambda_{Q} Q_j^T I) - \frac{D}{2}N_1 ln(\sigma_{U}^2)\nonumber\\
& -\frac{D}{2}N_2 ln(\sigma_{V}^2)+ DN_1 ln(\lambda_{P})+DN_2 ln(\lambda_{Q}) \nonumber\\
& - \frac{1}{2} \sum_{i=1}^{N_1} \sum_{j=1}^{N_2}I_{ij} \ln(P_i^TQ_j+\Delta \sigma^2) + const, \nonumber\\
& \text{subject to   } P_{ik},Q_{jk} \geq 0, i \in [1,N_1],j \in [1,N_2].
\end{align}

The non-negative constraint is given under exponential distribution. 
Note that the minimization for the prediction error $(R_{ij} - U_i^TV_j)^2$ is controlled by the deviation $\bar{\sigma}^2_{ij}$, which is also learned through factorization $P_i^TQ_j$.

\subsection{Learning DMF}

To learn DMF, one plausible strategy is to resort to stochastic gradient ascent (SG) for maximizing the log posterior distribution (MAP) since the objective function is differentiable. However, with the non-negative constraint on deviation latent factors, learning $P$ and $Q$ is not trivial. The constraint results in non-negative MF with the deviation latent factors. Fortunately, there have been several methods proposed for such constraint \cite{lin2007projected}, while we exploit a simple yet effective method called the projected gradient method. That is, every time a single step SG is done, all negative results in deviation latent factors are replaced with 0. 

It is not hard to realize that the SG learning of DMF is linear to the number of observed data. That is, O(\# of epochs $\times$ \# of observed data $\times$ D). Moreover, the space complexity is O(($N_1+N_2$) $\times$ D). Both time and space complexity are the same as conventional MF model. In our experiment, the deviation latent factors achieve at least $33\%$ sparsity through exponential prior. Finally, DMF can be considered as a joint learning machine to learn both mean and deviation, whose advantage will be discussed in the next section.



\begin{table*}[!t]
\small
\renewcommand{\arraystretch}{1.3}
\caption{Statistics of recommender system (RS) and Internet of Things (IoT) Datasets}
\label{statistics}
\centering
\begin{tabular}{ c | c c c || c | c c }
\hline
RS & MovieLens1M & MovieLens10M & Netflix & IoT & London & Berkeley\\
\hline
 \# of users & 6,040 & 69,878 & 480,189 & \# of locations & 15 & 58\\
\# of movies & 3,706 & 10,677 & 17,770 & \# of sensors & 19 & 4\\
- & - & - & - & \# of timestamps & 7,479 & 46,613\\
\# of data &  1M & 10M & 100M & \# of data & 476,638 & 9,158,515\\
sparsity & 95.53\% & 98.66\% & 98.82\% & sparsity & 77.64\% & 15.31\%\\
\hline
\end{tabular}
\end{table*}

\section{Discussion}
In this section, we provide some insights of deviation-driven learning to distinguish its roles in the learning process.

\subsection{Deviation as Learning Weight Scheduling}
In literature, several methods have been proposed for the per-coordinate learning rate scheduling on first order gradient descent learning \cite{duchi2011adaptive,riedmiller1993direct,zeiler2012adadelta}. For example, the AdaGrad optimizer \cite{duchi2011adaptive} conducts learning rate scheduling for a coordinate $x$ by scaling the current learning rate by the square root of sum of all squared historical gradients of $x$.

Our model treats weight scheduling from different angle. Instead of scheduling on different coordinates, i.e., $x=U_{ik},V_{jk}$, in a model, our model essentially provides dynamic learning weights on the cost $(R_{ij}-\bar{R}_{ij})^2$ for different instances $R_{ij}$, during optimization, as shown in (\ref{lrefor_object}). Besides, the weight scheduling is not conducted on the regularization for parameters, i.e., the priors; although a noisy data point should receive a lower cost on the training, the regularization of corresponding parameters should not be equally down-weighted during training.



There are some orthogonal works in weighting data in literature. For instance, \cite{bengio2009curriculum,kumar2010self} propose to organize, rather than randomly present as SG, data points in a meaningful manner for training a better non-convex model. Since the proposed deviation-driven modeling does not influence the optimization process, it can be coordinated with \cite{bengio2009curriculum,kumar2010self}. Similarly, our method can also be easily combined with the conventional per-coordinate learning rate scheduling strategy. 

\subsection{Deviation as Dynamic Regularization}

Deviation can also be viewed as a regularization scheme for different instances. As deviation controls the fitting for data, it yields a looser constraint of fitting for less informative instance with higher deviation, and vice versa. Accordingly, faster convergence can be expected as it does not need to overfit the noisy data (i.e., those with high deviation). Besides, contrary to the conventional regularization scheme that manipulates regularization on model parameters to control the fitting on data, deviation-driven learning directly regularize the fitting for noisy data by deviation estimation.

Also note that the deviation of our model is learned jointly with the main prediction function, so deviation can dynamically coordinate with main prediction throughout the whole optimization process. Thus, in each iteration, the deviation can be adjusted and immediately used to re-weight the cost function to learn a better prediction model. In contrast, in traditional weighting method \cite{koren2009matrix}, the confidence level of data point is assumed to be explicitly given, or learned separately with the model. 

\subsection{Deviation Model as a General Methodology}

Since the modeling of deviation involves the assumption of the existence of ubiquitous and diverse noise, it is worthy to consider when to adopt deviation-driven learning. On one hand, if the noises implicitly or explicitly exist, it is clear our model is useful. On the other hand, given perfectly clean data, our model also allows a solution with shared/near zero deviation value. As a result, given insufficient knowledge about whether data are noisy, DMF is a better choice than typical Biased MF, which is validated in the next section.



\section{Experiments}

To validate the efficacy of deviation-modeling, we conduct experiments on both \emph{noisy} Internet of Things (IoT) and \emph{clean} recommender system (RS) datasets. Especially, superior performance of noise modeling in the "clean" datasets may indicate the existence of implicit noise as we hypothesized.


\subsection{DMF for Recommendation}

Experiments are conducted on three standard movie recommendation datasets: MovieLens1M, MovieLens10M, and Netflix dataset. Each dataset can be represented by a matrix $R$: the two dimensions denote user and movie, respectively. The statistics is shown in the left of Table \ref{statistics}.

The experiments are conducted using the protocol proposed in the state-of-the-art solutions \cite{lee2013local,sedhain2015autorec}. Specifically, for each dataset, we produce a random 90\% and 10\% split of data for training and testing, and repeat 5 times to generate 5 different splits. Among the training data, 10\% of them are used as validation set for parameter tuning. For testing data without user or movie observed in training data, a default rating 3 is predicted by convention \cite{sedhain2015autorec}. We run our experiments in each split and report the average root mean squared error (RMSE) on testing data.


We compare DMF with following collaborative filtering (CF) methods:
\begin{itemize}
  \item \emph{Restricted Boltzmann Machines (RBM)} \cite{chen2011linear}. A widely used generative model for CF.
  \item \emph{Biased Matrix Factorization (Biased MF)}. A simple yet successful method for recommendation as introduced in (\ref{b_MF}), the non-deviation-driven counterpart of DMF. 
  \item \emph{Robust Collaborative Filtering (RCF)} \cite{mehta2007robust}. A robust weighted MF model applying Huber M-estimator \cite{huber2011robust} to iteratively estimate the weighting during optimization. $L_2$ regularization is added for better generalization performance.
  \item \emph{Robust Bayesian Matrix Factorization (RBMF)} \cite{lakshminarayanan2011robust}. A robust MF adopting a rank-1 structure for noise and Student-t prior on mean latent factors.
  \item \emph{Local Low-Rank Matrix Approximation (LLORMA)} \cite{lee2013local}. The state-of-the-art matrix approximation model assuming the target matrix is locally low-rank.
  \item \emph{AutoRec} \cite{sedhain2015autorec}. The state-of-the-art deep learning recommendation model based on Auto-Encoder.
\end{itemize}
We also conduct experiments for the state-of-the-art anomaly detection method, Direct Robust MF (DRMF) \cite{xiong2011direct}, which imposes an upper-bound on the cardinality of detected noise set to guarantee the sparsity of noise. Since DRMF is designed for anomaly detection in full matrix, Singular Value Decomposition is used for factorization. We make a generalization for CF by adopting Biased MF for factorization. \{10\%, 5\%, 1\%\} of the training data as the upper-bound are tested in MovieLens1M and MovieLens10M data, but in all the setting the training and testing performance of DRMF decrease at almost every iteration, thus is not suitable for comparison.


AdaGrad learning rate scheduling \cite{duchi2011adaptive} is used for training DMF. Parameters are fine-tuned through validation. Detail parameters are omitted due to space limitation. The complete parameter setting and implementation can be found at \footnote{wait until camera-ready}.

The experiment results are presented in Table \ref{rmse_rec}. DMF outperforms all the state-of-the-art in MovieLens10M and Netflix by a large margin. The only exception comes from the smaller dataset (i.e., MovieLens1M) where our model might have overfitted the training data due to insufficient data. 

Since DMF, Biased MF, RCF, and RBMF share similar search space for main prediction, it is worthy to compare the these MF models. First, DMF outperforms Biased MF, its non-deviation-driven counterpart, significantly in all datasets. Since the three datasets are not recognized as noisy in literature, the phenomenon may imply potential implicit noise on the data as we hypothesized. It demonstrates that jointly learning the deviation and rating yields more accurate prediction. For RCF, using an M-estimator to estimate the weight is even worse than vanilla Biased MF. Compared with RBMF, DMF is superior in larger datasets, demonstrating the efficacy of low-rank (e.g., 100) variance structure versus the rank-1 variance structure in RBMF.

Finally, compared with nonlinear AutoRec model, the superior performance of linear DMF is observed. We argue the phenomenon comes from the sparsity of real world data that a nonlinear model is prone to overfit. Similar scenario appears in knowledge base completion that a linear model is reported to dominate non-linear models \cite{yang2014embedding}.

\begin{table}[!t]
\small
\renewcommand{\arraystretch}{1.3}
\caption{Performance in RS: all RMSE values have 95\% confidence intervals $\leq$ $\pm$0.003 except RCF ($\pm$0.005).}
\label{rmse_rec}
\centering
\begin{tabular}{c c c c}
\hline
{ } & MovieLens1M & MovieLens10M & Netflix\\
\hline
RBM & 0.881 & 0.823 & 0.845\\
Biased MF & 0.845 & 0.803 & 0.844\\
RCF & 0.867 & 0.809 & 0.855 \\
RBMF & 0.837 & 0.795 & 0.824\\
LLORMA & 0.833 & 0.782 & 0.834\\
AutoRec & \textbf{0.831} & 0.782 & 0.823\\
DMF & 0.841 & \textbf{0.775} & \textbf{0.806}\\
\hline
\end{tabular}
\end{table}

\subsection{Missing Sensor Data Imputation on IoT data}

The task of missing sensor data imputation is to infer missing testing data given observed training data. The experiments are conducted on two real world IoT datasets, an outdoor air quality monitoring dataset from London and an indoor sensor dataset from Intel Berkeley Research lab, which is open to public\footnote{http://db.csail.mit.edu/labdata/labdata.html}. The IoT data are naturally noisy due to sensor measurement errors. Each dataset can be represented by a 3-mode tensor $T$: the three dimensions denote location, timestamp, and sensor type, respectively. Note that there are multiple types of sensors (e.g., humidity, light and heat sensors) in each location. The statistics are shown in the right of Table \ref{statistics}. Since multi-modal sensors may have different ranges of values, data are normalized for each type of sensor by standard score. 

We make a generalization of DMF to Deviation-driven Tensor Factorization (DTF) based on PARAFAC tensor factorization \cite{xiong2010temporal} for modeling mean and variance, respectively. The time complexity of DTF is trivially 1.5 times of DMF. All experiments are repeated 100 times, and the average result is reported. Here we exploit mean squared error (MSE) as our evaluation metric. DTF is compared with the following classic sensor data imputation and factorization models:

\begin{figure}
\begin{subfigure}{.214\textwidth}
  \centering
  \includegraphics[width=1\linewidth]{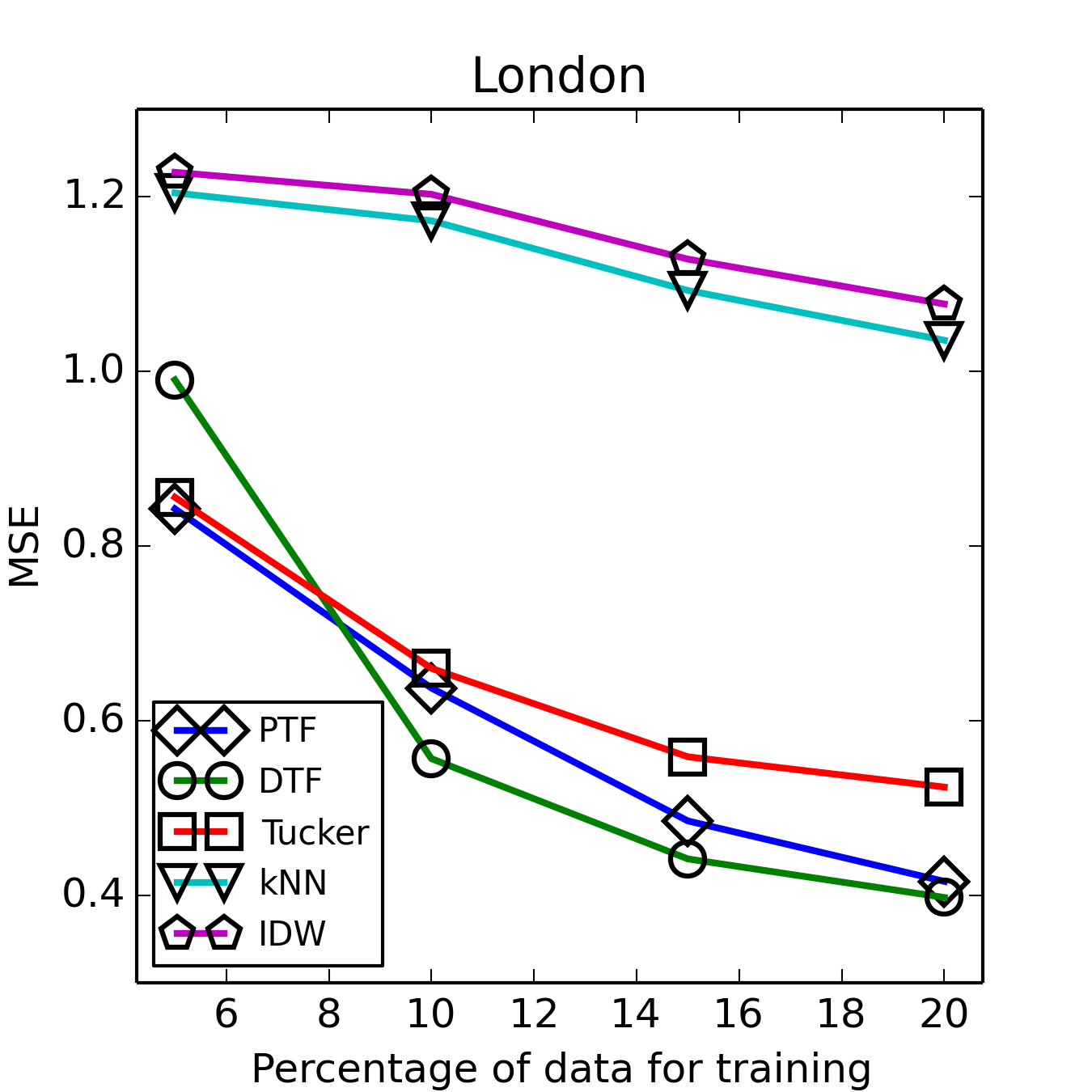}
\end{subfigure}%
\begin{subfigure}{.285\textwidth}
  \centering
  \includegraphics[width=1\linewidth]{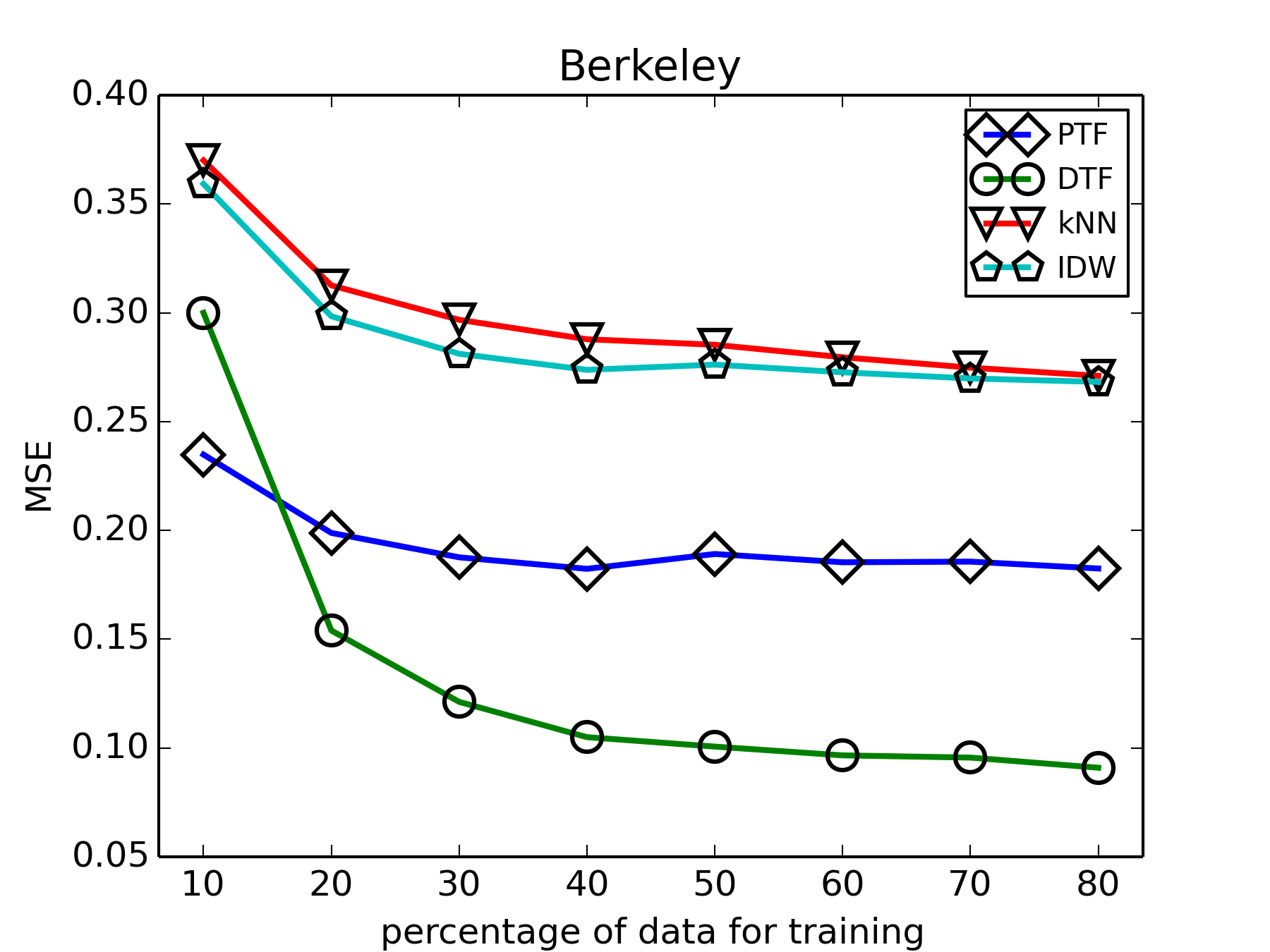}
\end{subfigure}
\caption{Prediction error over various percentage of training data in London and Berkeley IoT datasets. The range of x-axis may vary according to the original sparsity of dataset. For example, we only have $22.36\%$ of data available in London data, so we conduct experiments using $\{5\%,10\%,15\%,20\%\}$ of training data. 
}
\label{fig:iot_mse}
\end{figure}

\begin{itemize}
  \item \emph{Spatial k-Nearest Neighbors (kNN)}. Average of $K$ spatially nearest data points as prediction. 
  \item \emph{Inversed Distance Weighting (IDW)}. Using the inverse of distance for data points as the weighting for interpolation.
  \item \emph{Probabilistic Tensor Factorization (PTF)} \cite{xiong2010temporal}. PTF is equivalent to PARAFAC tensor factorization with $L_2$ norm regularization, the non-deviation-driven counterpart of DTF.
  \item \emph{Tucker} \cite{kolda2009tensor}. Tucker Decomposition with $L_2$ regularization added for better performance.
\end{itemize}


The experiment are conducted through various percentages of data as training data. The setting for each dataset may vary according to its original sparsity. Results are shown in Figure \ref{fig:iot_mse}. The result of Tucker is not shown in Berkeley dataset because it is significantly worse than the others. It can be noticed that except the most sparse setting in each dataset, our method outperforms all competitors significantly. A reasonable explanation is that when data are already sparse, the downgrading of certain potentially noisy data can lead to insufficient amount of information for training. For winning cases, we also conduct unpaired t-test and confirmed that all P values are smaller than 0.0001. In summary, this section demonstrates the extendibility of DMF to tensor factorization model and the efficacy of deviation-driven learning on noisy data.

\subsection{Running Time and Convergence}

\begin{table}[!t]
\small
\renewcommand{\arraystretch}{1.3}
\caption{Running time analysis (in seconds) in MovieLens1M datasets}
\label{ml1m_time}
\centering
\begin{tabular}{c c c c}
\hline
{ } & Biased MF & AutoRec & DMF\\
\hline 
Total running time & 207 & 647 & 88\\
Running time/epoch & 2.1 & 3.3 & 4.4\\
\hline
\end{tabular}
\end{table}

We also compare the running time with the strongest competitor, AutoRec, and the counterpart of DMF, Biased MF, in MovieLens1M. 
All above models are implemented in C++. AdaGrad optimizer \cite{duchi2011adaptive} is used for DMF and Biased MF, and resilient propagation optimizer \cite{riedmiller1993direct} is used for AutoRec, as suggested in the original paper \cite{sedhain2015autorec}. 
The running time for each model is shown in Table \ref{ml1m_time}. Though the running time per epoch of DMF is the longest, it takes much fewer epochs to converge and in turn spent the least amount of total running time.

\begin{figure}
\centering
  \includegraphics[width=1\linewidth]{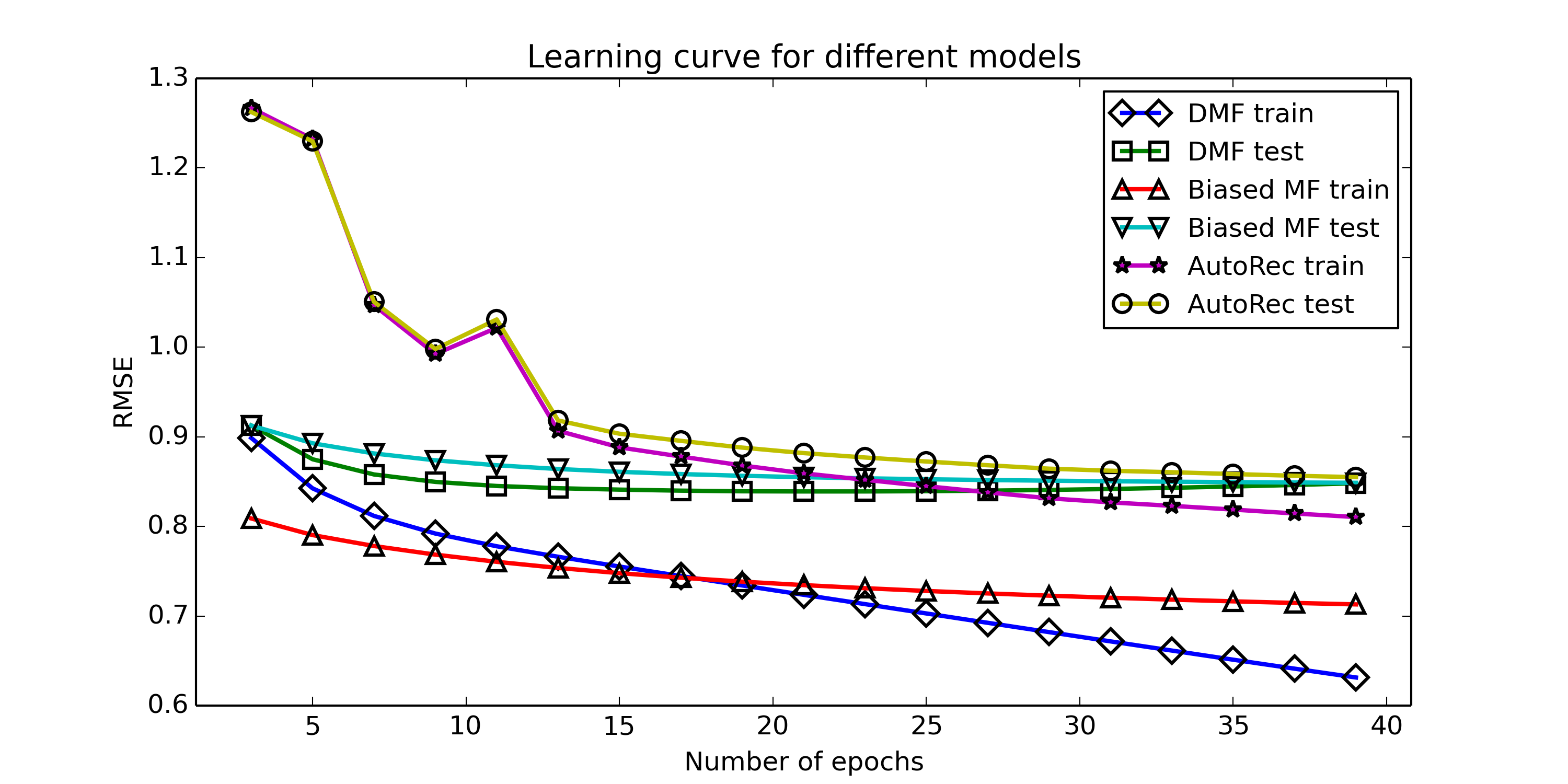}
  \caption{Learning curves in MovieLens1M.}\label{fig:lr2}
\end{figure}

We also conduct learning curve comparison of DMF with Biased MF and AutoRec. The results are shown in Figure \ref{fig:lr2}. Comparing Biased MF with DMF, 
it confirms our hypothesis that by downplaying the importance of data with larger deviation (i.e., noisy data), our model can converge faster than Biased MF. 
In summary, DMF achieves its best result in the shortest running time and number of epochs. 


\section{Related Works}

There are some related works regarding weighted MF, robust MF, and the structure of noise modeling. 

In traditional weighting method \cite{koren2009matrix}, the confidence level of data point is assumed to be explicitly given, or learned separately with the model. \cite{hu2008collaborative} and \cite{liang2016modeling} are designed for an implicit feedback scenario by a weighting mechanism, while our method addresses an explicit feedback scenario. \cite{schnabel2016recommendations} aims at weighting the distribution of data to get unbiased estimation, but we argue that real world data, like MovieLens1M and Netflix, are biased and thus should use biased data for evaluation. 

Robust MF is a family of MF models addressing noise modeling, but lots of works are designed for full matrix factorization rather than CF. For example, \cite{meng2013robust} and \cite{cao2015low} aim at recovering clean image. For robust CF, RCF \cite{mehta2007robust} proposes to use Huber M-estimator \cite{huber2011robust} to estimate the weighting, compared with our learning strategy. RBMF \cite{lakshminarayanan2011robust} is the most related one, which also models heteroscedastic variance; however, RBMF adopts a rank-1 variance structure, rather than our low-rank structure, in order to constitute Student-t prior on mean latent factor. In addition, the formulation of RBMF is intractable, requiring variational inference method to derive a tractable lower-bound, while DMF can be trained directly by SG. In our empirical study, RCF and RBMF are inferior to DMF in large datasets.

In literature, a great portion of noise-aware models adopt a sparse structure for noise \cite{xiong2011direct}, contrary to our low-rank structure. The discrepancy stems from the underlying assumption. A sparse noise structure deals with only a few (salient) noises, while we assume noise is everywhere. Moreover, low-rank structure is an efficient methodology to avoid overfitting the noises.

\section{Conclusions}
This paper proposes a deviation-driven objective function that considers diverse weights for each instance while learning a model. The main challenge lies in the fact that the deviation, or noise level, of each instance is unknown, which leads to the major contribution of this paper to propose a joint learning framework that can dynamically estimate the deviation and utilize it to adjust the cost function during training. Note that the recommendation datasets we used are all not known as noisy, which implies that even when data are clean, modeling the deviation as proposed can still yield superior performance, maybe due to the fact that data are actually implicitly noisy or not equally important in model training. Future works include exploitation of the idea to other learning tasks such as classification and clustering.

\bibliographystyle{aaai} 
\bibliography{output.bbl}

\begin{thebibliography}{}

\bibitem[\protect\citeauthoryear{Agarwal and Chen}{2010}]{agarwal2010flda}
Agarwal, D., and Chen, B.-C.
\newblock 2010.
\newblock flda: matrix factorization through latent dirichlet allocation.
\newblock In {\em Proceedings of the third ACM international conference on Web
  search and data mining},  91--100.
\newblock ACM.

\bibitem[\protect\citeauthoryear{Bengio \bgroup et al\mbox.\egroup
  }{2009}]{bengio2009curriculum}
Bengio, Y.; Louradour, J.; Collobert, R.; and Weston, J.
\newblock 2009.
\newblock Curriculum learning.
\newblock In {\em Proceedings of the 26th annual international conference on
  machine learning},  41--48.
\newblock ACM.

\bibitem[\protect\citeauthoryear{Cao \bgroup et al\mbox.\egroup
  }{2015}]{cao2015low}
Cao, X.; Chen, Y.; Zhao, Q.; Meng, D.; Wang, Y.; Wang, D.; and Xu, Z.
\newblock 2015.
\newblock Low-rank matrix factorization under general mixture noise
  distributions.
\newblock In {\em Proceedings of the IEEE International Conference on Computer
  Vision},  1493--1501.

\bibitem[\protect\citeauthoryear{Chen \bgroup et al\mbox.\egroup
  }{2011}]{chen2011linear}
Chen, P.-L.; Tsai, C.-T.; Chen, Y.-N.; Chou, K.-C.; Li, C.-L.; Tsai, C.-H.; Wu,
  K.-W.; Chou, Y.-C.; Li, C.-Y.; Lin, W.-S.; et~al.
\newblock 2011.
\newblock A linear ensemble of individual and blended models for music rating
  prediction.
\newblock {\em KDDCup}.

\bibitem[\protect\citeauthoryear{Duchi, Hazan, and
  Singer}{2011}]{duchi2011adaptive}
Duchi, J.; Hazan, E.; and Singer, Y.
\newblock 2011.
\newblock Adaptive subgradient methods for online learning and stochastic
  optimization.
\newblock {\em The Journal of Machine Learning Research} 12:2121--2159.

\bibitem[\protect\citeauthoryear{Gopalan, Hofman, and
  Blei}{2013}]{gopalan2013scalable}
Gopalan, P.; Hofman, J.~M.; and Blei, D.~M.
\newblock 2013.
\newblock Scalable recommendation with poisson factorization.
\newblock {\em arXiv preprint arXiv:1311.1704}.

\bibitem[\protect\citeauthoryear{Hu, Koren, and
  Volinsky}{2008}]{hu2008collaborative}
Hu, Y.; Koren, Y.; and Volinsky, C.
\newblock 2008.
\newblock Collaborative filtering for implicit feedback datasets.
\newblock In {\em 2008 Eighth IEEE International Conference on Data Mining},
  263--272.
\newblock Ieee.

\bibitem[\protect\citeauthoryear{Huber}{2011}]{huber2011robust}
Huber, P.~J.
\newblock 2011.
\newblock {\em Robust statistics}.
\newblock Springer.

\bibitem[\protect\citeauthoryear{Kolda and Bader}{2009}]{kolda2009tensor}
Kolda, T.~G., and Bader, B.~W.
\newblock 2009.
\newblock Tensor decompositions and applications.
\newblock {\em SIAM review} 51(3):455--500.

\bibitem[\protect\citeauthoryear{Koren, Bell, and
  Volinsky}{2009}]{koren2009matrix}
Koren, Y.; Bell, R.; and Volinsky, C.
\newblock 2009.
\newblock Matrix factorization techniques for recommender systems.
\newblock {\em Computer} (8):30--37.

\bibitem[\protect\citeauthoryear{Kumar, Packer, and
  Koller}{2010}]{kumar2010self}
Kumar, M.~P.; Packer, B.; and Koller, D.
\newblock 2010.
\newblock Self-paced learning for latent variable models.
\newblock In {\em Advances in Neural Information Processing Systems},
  1189--1197.

\bibitem[\protect\citeauthoryear{Lakshminarayanan, Bouchard, and
  Archambeau}{2011}]{lakshminarayanan2011robust}
Lakshminarayanan, B.; Bouchard, G.; and Archambeau, C.
\newblock 2011.
\newblock Robust bayesian matrix factorisation.
\newblock In {\em AISTATS},  425--433.

\bibitem[\protect\citeauthoryear{Lee \bgroup et al\mbox.\egroup
  }{2013}]{lee2013local}
Lee, J.; Kim, S.; Lebanon, G.; and Singer, Y.
\newblock 2013.
\newblock Local low-rank matrix approximation.
\newblock In {\em Proceedings of The 30th International Conference on Machine
  Learning},  82--90.

\bibitem[\protect\citeauthoryear{Liang \bgroup et al\mbox.\egroup
  }{2016}]{liang2016modeling}
Liang, D.; Charlin, L.; McInerney, J.; and Blei, D.~M.
\newblock 2016.
\newblock Modeling user exposure in recommendation.
\newblock In {\em Proceedings of the 25th International Conference on World
  Wide Web},  951--961.
\newblock International World Wide Web Conferences Steering Committee.

\bibitem[\protect\citeauthoryear{Lin}{2007}]{lin2007projected}
Lin, C.-J.
\newblock 2007.
\newblock Projected gradient methods for nonnegative matrix factorization.
\newblock {\em Neural computation} 19(10):2756--2779.

\bibitem[\protect\citeauthoryear{Mehta, Hofmann, and
  Nejdl}{2007}]{mehta2007robust}
Mehta, B.; Hofmann, T.; and Nejdl, W.
\newblock 2007.
\newblock Robust collaborative filtering.
\newblock In {\em Proceedings of the 2007 ACM conference on Recommender
  systems},  49--56.
\newblock ACM.

\bibitem[\protect\citeauthoryear{Meng and De~La~Torre}{2013}]{meng2013robust}
Meng, D., and De~La~Torre, F.
\newblock 2013.
\newblock Robust matrix factorization with unknown noise.
\newblock In {\em Proceedings of the IEEE International Conference on Computer
  Vision},  1337--1344.

\bibitem[\protect\citeauthoryear{Mnih and
  Salakhutdinov}{2007}]{mnih2007probabilistic}
Mnih, A., and Salakhutdinov, R.
\newblock 2007.
\newblock Probabilistic matrix factorization.
\newblock In {\em Advances in neural information processing systems},
  1257--1264.

\bibitem[\protect\citeauthoryear{Riedmiller and
  Braun}{1993}]{riedmiller1993direct}
Riedmiller, M., and Braun, H.
\newblock 1993.
\newblock A direct adaptive method for faster backpropagation learning: The
  rprop algorithm.
\newblock In {\em Neural Networks, 1993., IEEE International Conference on},
  586--591.
\newblock IEEE.

\bibitem[\protect\citeauthoryear{Salakhutdinov and
  Mnih}{2008}]{salakhutdinov2008bayesian}
Salakhutdinov, R., and Mnih, A.
\newblock 2008.
\newblock Bayesian probabilistic matrix factorization using markov chain monte
  carlo.
\newblock In {\em Proceedings of the 25th international conference on Machine
  learning},  880--887.
\newblock ACM.

\bibitem[\protect\citeauthoryear{Schnabel \bgroup et al\mbox.\egroup
  }{2016}]{schnabel2016recommendations}
Schnabel, T.; Swaminathan, A.; Singh, A.; Chandak, N.; and Joachims, T.
\newblock 2016.
\newblock Recommendations as treatments: Debiasing learning and evaluation.
\newblock {\em arXiv preprint arXiv:1602.05352}.

\bibitem[\protect\citeauthoryear{Sedhain \bgroup et al\mbox.\egroup
  }{2015}]{sedhain2015autorec}
Sedhain, S.; Menon, A.~K.; Sanner, S.; and Xie, L.
\newblock 2015.
\newblock Autorec: Autoencoders meet collaborative filtering.
\newblock In {\em Proceedings of the 24th International Conference on World
  Wide Web Companion},  111--112.
\newblock International World Wide Web Conferences Steering Committee.

\bibitem[\protect\citeauthoryear{Shalizi}{forthcoming}]{GAUSSMARKOV}
Shalizi, C.~R.
\newblock forthcoming.
\newblock {\em Advanced Data Analysis from an Elemntary Point of View}.
\newblock Cambridge, England: Cambridge University Press.

\bibitem[\protect\citeauthoryear{Xiong \bgroup et al\mbox.\egroup
  }{2010}]{xiong2010temporal}
Xiong, L.; Chen, X.; Huang, T.-K.; Schneider, J.~G.; and Carbonell, J.~G.
\newblock 2010.
\newblock Temporal collaborative filtering with bayesian probabilistic tensor
  factorization.
\newblock In {\em SDM}, volume~10,  211--222.
\newblock SIAM.

\bibitem[\protect\citeauthoryear{Xiong, Chen, and
  Schneider}{2011}]{xiong2011direct}
Xiong, L.; Chen, X.; and Schneider, J.
\newblock 2011.
\newblock Direct robust matrix factorizatoin for anomaly detection.
\newblock In {\em Data Mining (ICDM), 2011 IEEE 11th International Conference
  on},  844--853.
\newblock IEEE.

\bibitem[\protect\citeauthoryear{Yang \bgroup et al\mbox.\egroup
  }{2014}]{yang2014embedding}
Yang, B.; Yih, W.-t.; He, X.; Gao, J.; and Deng, L.
\newblock 2014.
\newblock Embedding entities and relations for learning and inference in
  knowledge bases.
\newblock {\em arXiv preprint arXiv:1412.6575}.

\bibitem[\protect\citeauthoryear{Zeiler}{2012}]{zeiler2012adadelta}
Zeiler, M.~D.
\newblock 2012.
\newblock Adadelta: An adaptive learning rate method.
\newblock {\em arXiv preprint arXiv:1212.5701}.

\end{thebibliography}

\end{document}